\documentclass[sn-mathphys,Numbered]{sn-jnl}

\usepackage{graphicx}%
\usepackage{graphicx}
\usepackage{multirow}%
\usepackage{pdflscape}
\usepackage{booktabs}
\usepackage{caption}
\usepackage{array}
\usepackage{amsmath,amssymb,amsfonts}%
\usepackage{amsthm}%
\usepackage{mathrsfs}%
\usepackage[title]{appendix}%
\usepackage{xcolor}%
\usepackage{soul}
\usepackage{textcomp}%
\usepackage{manyfoot}%
\usepackage{booktabs}%
\usepackage{algorithm}%
\usepackage{algorithmicx}%
\usepackage{algpseudocode}%
\usepackage{listings}%
\usepackage{adjustbox}
\usepackage[T1]{fontenc}

\theoremstyle{thmstyleone}%
\theoremstyle{thmstyletwo}%

\theoremstyle{thmstylethree}%

\makeatletter
\providecommand{\orcidID}[1]{}
\providecommand{\orcidlink}[1]{}
\makeatother

\begin{document}

\title[Beyond the star rating]{Beyond the Star Rating: A Scalable Framework for Aspect-Based Sentiment Analysis Using LLMs and Text Classification}

\author{Vishal Patil}\email{vspatil@usc.edu}

\author{Shree Vaishnavi Bacha}\email{sbacha@usc.edu}

\author{Revanth Yamani}\email{yamani@usc.edu}

\author{\fnm{Yidan} \sur{Sun}}\email{yidans@isi.edu}

\author*{\fnm{Mayank} \sur{Kejriwal}}\email{kejriwal@isi.edu}

\affil{\orgname{University of Southern California}, \orgaddress{\city{Marina del Rey}, \postcode{90292}, \state{CA}, \country{USA}}}



\abstract{Customer-provided reviews have become an important source of information for business owners and other customers alike. However, effectively analyzing millions of unstructured reviews remains challenging. While large language models (LLMs) show promise for natural language understanding, their application to large-scale review analysis has been limited by computational costs and scalability concerns. This study proposes a hybrid approach that uses LLMs for aspect identification while employing classic machine-learning methods for sentiment classification at scale. Using ChatGPT to analyze sampled restaurant reviews, we identified key aspects of dining experiences and developed sentiment classifiers using human-labeled reviews, which we subsequently applied to 4.7 million reviews collected over 17 years from a major online platform. Regression analysis reveals that our machine-labeled aspects significantly explain variance in overall restaurant ratings across different aspects of dining experiences, cuisines, and geographical regions. Our findings demonstrate that combining LLMs with traditional machine learning approaches can effectively automate aspect-based sentiment analysis of large-scale customer feedback, suggesting a practical framework for both researchers and practitioners in the hospitality industry and potentially, other service sectors.}

\keywords{Yelp, multi-factor, aspect-based analysis, sentiment analysis, regression models, text classification, large language model}



\maketitle

\section{Introduction}\label{intro}

Online review platforms have become influential in informing consumer decisions and impacting business outcomes, especially in the hospitality industry \cite{xie2014business, alalwan2017social, chevalier2006effect}. On platforms like Google Reviews, Yelp, and TripAdvisor, customers can now easily share their experiences with a broader audience through electronic word-of-mouth. Such platforms offer a structured way of publishing and disseminating reviews on official businesses \cite{schuckert2015hospitality}, in contrast with `organic' reviewing prevalent on social media channels such as Instagram and TikTok. The reliance on reviews-specific platforms has grown even more since the COVID-19 pandemic, with surveys indicating that the proportion of consumers who `always' or `regularly' read online reviews significantly increased, rising from 60\% in 2020 to 75\% in 2024 \cite{brightlocal2024}. This growing dependence on online reviews has made maintaining a positive reputation and responding to customer feedback essential for long-term business success. For instance, a study found that a one-star increase in Yelp rating leads to a 5–9\% increase in revenue  \cite{luca2016reviews}. Unsurprisingly, restaurateurs and hospitality managers have now come to rely heavily on these platforms to gather feedback and engage with customers \cite{kwok2017thematic}.

\begin{figure}[h]
    \centering
    \includegraphics[width=\textwidth]{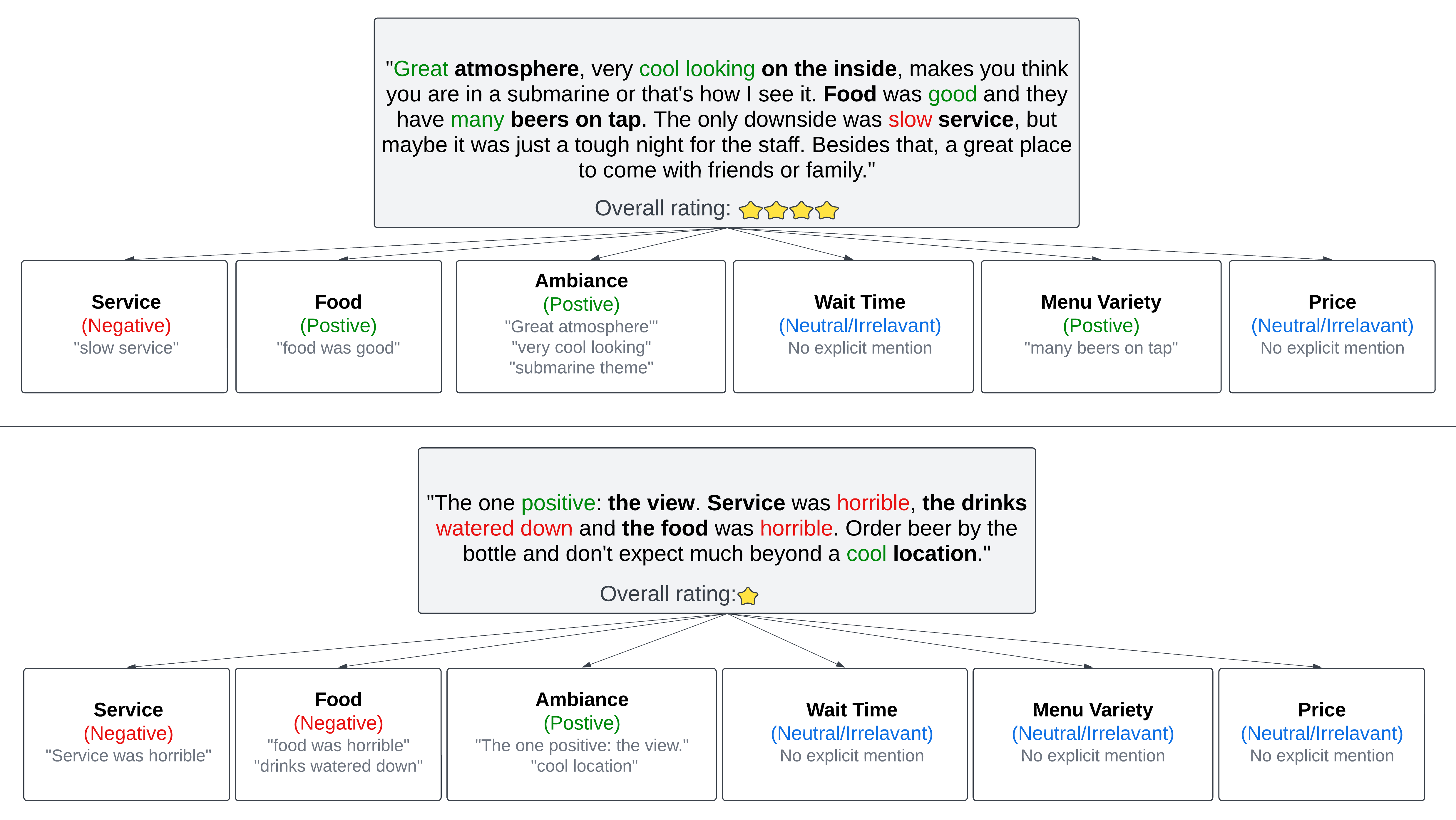} 
    \caption{Example of aspect-based sentiment analysis on two restaurant reviews. Identified aspects in the review text, such as \textbf{service}, \textbf{food}, and \textbf{ambiance}, are highlighted in bold. Positive sentiments are highlighted in green, and negative sentiments in red.}

    \label{Example}
\end{figure}

The high volume of data on such platforms also makes them amenable to automated analysis \cite{hu2004mining}. Researchers have extensively examined the relationship between domain-relevant attributes (e.g., service, food quality) and their association with customer satisfaction in the hospitality industry, initially using frequency-based methods \cite{xie2014business, pantelidis2010electronic} and later, machine learning approaches 
 \cite{zhao2019predicting}. Such domain-relevant attributes, called \textit{aspects}, may only be implicitly present in reviews. Also, customers may view some aspects neutrally, while having significantly negative (or positive) sentiments on others. Figure \ref{Example} illustrates two restaurant reviews, showing specific dining aspects with associated sentiments. Rather than conveying a single overall sentiment, reviews often reflect mixed sentiments across aspects (e.g., positive ambiance but negative food or service in the second review).

 Earlier studies often relied on predefined aspects based on qualitative methods (such as grounded theory and thematic analysis \cite{corbin1990grounded,braun2006using,guest2011applied}) to explore associations between various qualitative factors and overall customer satisfaction. However, manually analyzing large sets of unstructured and continuously updating review data is intractable, which motivates the need for automated aspect-mining and for extracting per-review sentiments associated with each mined aspect. This problem is denoted as \textit{Aspect-Based Sentiment Analysis (ABSA)}. Prior work has explored using topic modeling techniques such as Latent Semantic Allocation (LSA) 
 \cite{blei2003latent} to automatically discover the aspects evaluated in reviews and measure how sentiments toward different aspects are expressed 
 \cite{jo2011aspect, xu2016antecedents}. In recent years, deep learning methods, especially transformer-based models, have yielded significant progress on this problem of 
 \cite{schouten2015survey}. More complicated tasks, such as simultaneously identifying aspects, the opinions expressing the sentiment toward those aspects, and their associated sentiment polarity, have also been studied \cite{binder2019explaining, do2019deep}.

Despite these advances, obtaining good performance across large, unstructured sets of reviews remains a challenge, particularly when models need to adapt across different domains and geographies. Therefore, one interesting question is whether large language models (LLMs) can supplement aspect discovery and sentiment analysis. LLMs have shown promise in automating such tasks owing to background knowledge acquired through their massive pre-training, with applications in fields like law and healthcare \cite{drapal2023using, mathis2024inductive}. However, given the large corpora of real reviews collected across many different U.S. states over a considerable length of time, there are open questions around using LLMs directly for ABSA at scale. One challenge with using commercial LLMs is cost: prompting ChatGPT on millions of reviews is expensive, but may not be functionally necessary if cheaper machine learning methods can achieve similar performance without incurring high cost. We begin our study by evaluating the capability of a commercial LLM like ChatGPT \cite{openai2024chatgpt} to automate ABSA in large-scale restaurant review datasets.  We explore the premise that ChatGPT may be more useful for augmenting or replacing thematic analysis (identification of specific aspects), while sentiments on identified aspects could be reasonably predicted using off-the-shelf, locally deployable text classification tools that don't require LLMs to be prompted millions of times. 

Because it involves both LLMs and local methods, any such architecture for ABSA is complex, involving several options for models, parameters and other `design choices.' Our first methodological contribution is to demonstrate and evaluate potential design choices for a scalable, high-quality and cost-effective ABSA architecture that could be used for systematically coding the corpus of reviews for aspects and sentiments. Once designed, we use the architecture to process millions of reviews. We then demonstrate the collective utility of extracted aspects and their predicted sentiments for explaining the variance of \textit{overall} restaurant ratings. As shown in Figure \ref{Example}, these overall ratings (e.g., on a scale of 1-5) are provided directly by users. However, a systematic study at scale has thus far been lacking on quantitatively determining whether (and which) specific aspects and corresponding sentiments can \textit{explain} the variance in these ratings. 

With the background above in place, we state the following two goals:

\begin{enumerate} 
    \item \textbf{Aspect identification and sentiment labeling:} 
    We sample, identify, and categorize the key aspects from user reviews alone, using an LLM like ChatGPT, that could be potentially associated with actual user experiences. We also propose and validate scalable off-the-shelf text classification models to automatically extract sentiment labels in each review for each identified aspect. The final result is an ABSA architecture that makes judicious use of LLMs and local machine learning methods, and that we apply on millions of reviews from the Yelp platform. 
    
    \item \textbf{Impact analysis on restaurant ratings:} Using the predicted aspect-sentiments, we specify and fit a regression model to determine whether the predicted aspects and sentiments can explain the variance in overall restaurants' ratings. In other words, we aim to measure whether, despite the noise in the reviews and the implicit content in their text (where aspects and sentiments are usually only latent, if at all), aspects provide explanatory power for predicting why some restaurants are rated highly compared to others. The analysis is completely quantitative, and backed by rigorous statistical testing. Through further analysis of coefficients, we also aim to quantify the relative importance of different aspects and to assess the role, if any, that the noise in the automatic sentiment labeling process may have played in the final quality of the regression.
\end{enumerate}

We investigate these objectives by collecting and analyzing 4.7 million reviews published over a 17-year period from the Yelp online platform in the U.S. and Canada. Experimentally, we find that sentiment predictions based on ChatGPT-identified key aspects from customer reviews, which were trained on human annotations, can reliably account for much of the variance in reviewers' final aggregate ratings. Our approach offers a framework for analyzing large volumes of customer feedback in the hospitality industry, with potential applications in other service sectors.

\section{Materials and Methods}\label{methods}

\subsection{Data}

This study uses the Yelp Open Dataset \cite{yelp_dataset}, a publicly available representative sample of business, user, and review data provided by Yelp, spanning various metropolitan areas primarily in the United States and Canada. The dataset is organized into several JavaScript Object Notation (JSON) files, each representing a different entity (businesses, reviews, and users), which can be joined through unique identifiers.

For this analysis, we primarily utilized the `review' and `business' files. The review file contains individual reviews, including the review text, rating (on a scale of 1-5), and timestamp, with each review uniquely identified and linked to a specific business. The `business' file contains business information such as location, business categories, and other metadata, with each business uniquely identified. Since this study focuses exclusively on restaurant businesses, we first merged the two files using the business identifier as the common key, then filtered the merged dataset to include only those businesses where `restaurant' was listed in their `categories' field. 

The final dataset spans from 2005 to 2022 and consists of 4,724,684 restaurant reviews written by 1,446,031 unique users for 52,286 distinct restaurant businesses. The data covers 17 U.S. states and one Canadian province, with Pennsylvania having the highest number of reviews at 1,100,276, while Montana had the fewest, with only six reviews. Across all restaurants, the average overall business rating is 3.793, while the average individual review rating is 3.794. The variables used in this study are detailed in Table \ref{dataset}.

\begin{table}[h]
\centering
\begin{tabular}{|p{4cm}|p{9cm}|}
\hline
\multicolumn{2}{|c|}{\textbf{Fields Used in the Study}} \\ \hline
\textbf{Variable} & \textbf{Description} \\ \hline
\texttt{review\_id} & Unique identifier for each review \\ \hline
\texttt{user\_id} & Unique identifier for each user \\ \hline
\texttt{business\_id} & Unique identifier for each business \\ \hline
\texttt{review\_rating} & Star rating assigned by the user to the business (1-5 scale) \\ \hline
\texttt{text} & Full text of the review provided by the user \\ \hline
\texttt{date} & Date when the review was submitted \\ \hline
\texttt{name} & Name of the business \\ \hline
\texttt{address} & Full address of the business \\ \hline
\texttt{city} & City where the business is located \\ \hline
\texttt{state} & U.S. state or Canadian province where the business is located \\ \hline
\texttt{postal\_code} & Postal code of the business location \\ \hline
\texttt{overall\_rating} & Average overall rating of the business (1-5 scale) \\ \hline
\texttt{review\_count} & Total number of reviews the business has received \\ \hline
\texttt{categories} & List of categories or tags associated with the business (e.g., restaurant types, services offered) \\ \hline
\end{tabular}
\caption{Description of variables from the Yelp Open Dataset used for analysis.}
\label{dataset}
\end{table}

\subsection{Aspect Discovery and Inter-Annotator Consistency}
To identify recurring aspects of dining experiences in customer reviews, we randomly selected 30 restaurants in California and sampled 20 reviews from each. These 600 reviews were input into two models, ChatGPT-4o and ChatGPT-4o mini, with the prompt:

\textit{`From the customer reviews above, identify key aspects of the dining experience for labeling purposes. For each identified aspect, assign a rating to each review on a 5-point scale.'}

Table \ref{chatgpt} presents representative examples of the key aspects identified by both models across five sample businesses (with business IDs masked). We evaluated both occurrences (how often an aspect was identified by either model) and inter-model agreement (the percentage of businesses where both models identified the same aspect) for each aspect. The most frequent aspects were \textit{service} (100\% occurrence, 100\% agreement), \textit{food quality} (96.67\% occurrence, 93.10\% agreement), \textit{price} (93.33\% occurrence, 100\% agreement), \textit{atmosphere/ambiance} (86.67\% occurrence, 86.67\% agreement), \textit{menu variety} (66.67\% occurrence, 83.33\% agreement), and \textit{wait time} (63.33\% occurrence, 65\% agreement).  Note that some minor aspects were consolidated under broader categories. For example, \textit{boba quality} or \textit{sushi quality} were classified under \textit{food quality}. Less frequent or lower agreement aspects (<50\% occurrence or <60\% agreement) included \textit{portion sizes}, \textit{vegan options}, and several others.

\begin{table}[h]
    \centering
    \renewcommand{\arraystretch}{1.5} 
    \footnotesize
    \begin{tabular}{|c|p{10cm}|}
         \hline
         \multicolumn{2}{|c|}{\textbf{GPT-4o}} \\ \hline 
         \textbf{Bussiness ID} & \textbf{Extracted Aspects} \\ \hline
         \textbf{\texttt{cOoTI*******VZzdw}} & Service, Food Quality, Atmosphere/Ambiance, Value for Money, Speed of Service, Accuracy of Order, Drink Quality, Friendliness of Staff, Consistency, Special Dietary Offerings \\ \hline
         \textbf{\texttt{2N9WH*******-qN\_A}} & Food Quality, Service, Beverage Quality, Ambiance, Price/Value, Accuracy of Order, Consistency, Dietary Options, Wait time \\ \hline
         \textbf{\texttt{6EizP*******ZMI9Lw}} & Service, Food Quality, Ambiance, Presentation, Value for Money, Timeliness, Menu Variety, Consistency, Portion Size, Drink Selection \\ \hline
         \textbf{\texttt{cpaML*******\_GXow}} & Service, Boba Quality, Price, Sushi Quality, Location Convenience, Atmosphere/Ambiance, Portion Size, Authenticity, Customization Options \\ \hline
         \textbf{\texttt{RXA9e*******KHyIQ}} & Food Quality, Service, Atmosphere/Ambiance, Cleanliness, Portion Size, Value for Money, Speed/Timeliness, Authenticity, Consistency, Curb Appeal, Recommendations, Specialty Items \\ \hline
    \end{tabular}
    \vspace{0.5cm}
    \footnotesize 
    \begin{tabular}{|c|p{10cm}|}
         \hline
         \multicolumn{2}{|c|}{\textbf{GPT-4o-mini}} \\ \hline 
         \textbf{Bussiness ID} & \textbf{Extracted Aspects} \\ \hline
         \textbf{\texttt{cOoTI*******VZzdw}} & Food Quality, Service, Ambiance, Wait time, Value for Money, Menu Variety, Cleanliness, Staff Friendliness, Portion Size, Recommendation Likelihood \\ \hline
         \textbf{\texttt{2N9WH*******-qN\_A}} & Cupcake Quality, Service, Food Quality (general), Atmosphere/Ambiance, Coffee Quality, Value for Money, Selection Variety, Overall Satisfaction \\ \hline
         \textbf{\texttt{6EizP*******ZMI9Lw}} & Atmosphere/Ambiance, Food Quality, Service, Value for Money, Reservation Experience, Specific Dishes, Overall Experience \\ \hline
         \textbf{\texttt{cpaML*******\_GXow}} & Boba Quality, Sushi Quality, Service, Value for Money, Convenience, Variety of Menu Options, Customer Loyalty \\ \hline
         \textbf{\texttt{RXA9e*******KHyIQ}} & Food Quality, Cleanliness, Service, Atmosphere, Value for Money, Menu Variety, Takeout Experience, Outdoor Seating, Recommendation Likelihood \\ \hline
    \end{tabular}
    \caption{Key aspects extracted by GPT-4o and 4o-mini for five randomly selected businesses based on customer reviews. Business IDs are partially masked.}
    \label{chatgpt}
\end{table}

\begin{table}[h]
    \centering

    \begin{tabular}{|c|p{11cm}|}
        \hline
        \multicolumn{2}{|c|}{\textbf{Positive Sentiment Group}} \\ \hline
        \textbf{Topic 1} & great, food, place, service, restaurant, good, time, chicken, santa, chase \\ \hline
        \textbf{Topic 2} & best, clam, chowder, waiter, manhattan, italian, barbara, santa, favorite, great \\ \hline
        \textbf{Topic 3} & delicious, food, perfect, place, kitchen, great, love, cherry, little, half \\ \hline
        \textbf{Topic 4} & great, place, food, service, lunch, loved, drink, cheese, jonathan, good \\ \hline
        \textbf{Topic 5} & calamari, happy, hour, appetizer, great, drink, chicken, bartender, day, good \\ \hline
    \end{tabular}

    \vspace{0.5cm}

    \begin{tabular}{|c|p{11cm}|}
        \hline
        \multicolumn{2}{|c|}{\textbf{Negative Sentiment Group}} \\ \hline
        \textbf{Topic 1} & service, table, food, like, experience, place, completely, water, poached, came \\ \hline
        \textbf{Topic 2} & overpriced, customer, service, view, star, mgmt, zero, place, building, need \\ \hline
        \textbf{Topic 3} & dish, waiter, suggested, meat, salmon, bone, dry, fresh, specifically, fish \\ \hline
        \textbf{Topic 4} & steak, sauce, meal, good, great, view, probably, beach, window, best \\ \hline
        \textbf{Topic 5} & breakfast, drink, wait, free, hyatt, believe, restaurant, waited, pay, hotel \\ \hline
    \end{tabular}

    \vspace{0.5cm}

    \begin{tabular}{|c|p{11cm}|}
        \hline
        \multicolumn{2}{|c|}{\textbf{Neutral Sentiment Group}} \\ \hline
        \textbf{Topic 1} & ordered, good, sauce, salad, dessert, restaurant, asked, veal, really, cannoli \\ \hline
        \textbf{Topic 2} & fry, nice, price, plate, restaurant, way, lot, service, drop, soft \\ \hline
        \textbf{Topic 3} & food, waitress, restaurant, party, look, place, great, overall, pretty, chicken \\ \hline
        \textbf{Topic 4} & food, extra, sauce, charge, like, want, ask, minute, sat, place \\ \hline
        \textbf{Topic 5} & food, restaurant, service, place, table, great, good, time, really, like \\ \hline
    \end{tabular}
    
    \caption{Examples of the most relevant words in topics generated by Latent Dirichlet Allocation (LDA) across the three sentiment groups (positive, negative, neutral) for restaurant reviews. Each group consists of five topics, with the ten most frequently associated words shown for each topic.}
\label{lda}

\end{table}

\begin{table}[h!]
\centering
\begin{tabular}{|p{5cm}|c|c|c|c|c|c|}
\hline
\textbf{Review} & \textbf{Service} & \textbf{Food} & \textbf{Ambiance} & \textbf{Wait} & \textbf{Price} & \textbf{Menu} \\
& & \textbf{Quality} & & \textbf{Time} & & \textbf{Variety} \\ \hline
The New Mexican style cuisine here is damn good. I love the breakfast, especially with a bloody mary. Yum! The green and red sauces are delicious but SPICY so if you don't like spice consider the sour cream green chili sauce, or half and half. The staff are pretty great too. & 1 & 1 & NA & NA & NA & 1 \\ \hline
Great atmosphere, very cool looking on the inside, makes you think you are in a submarine or that's how I see it. Food was good and they have many beers on tap. The only downside was slow service, but maybe it was just a tough night for the staff. Besides that, a great place to come with friends or family. & -1 & 1 & 1 & NA & 0 & 1 \\ \hline
Really good fries{\ldots}about a dozen varieties from Spanish, Irish, Mexican, pizza \& more. Wings (their specialty) were great. Our waitress, Camilla, was quick, attentive, friendly \& helpful. Plus she was as cute as a button. Bathrooms were clean. 50\% off wings if you check in on Yelp!! We shall return. & 1 & 1 & 1 & 1 & 1 & 1 \\ \hline
The one positive: the view. Service was horrible, the drinks watered down and the food was horrible. Order beer by the bottle and don't expect much beyond a cool location. & -1 & -1 & 1 & NA & NA & NA \\ \hline
\end{tabular}
\caption{Example of sentiment labels for reviews assigned by one reviewer, with aspects rated by sentiment score: -1 (negative), 0 (neutral), 1 (positive), and `NA' for `irrelevant'.}
\label{labeled_data}
\end{table}

To validate these LLM-derived aspects, we conducted a comparative analysis using Latent Dirichlet Allocation (LDA)-based topic modeling \cite{blei2003latent}. LDA is a widely used probabilistic model for topic extraction from a large corpus of unstructured text. It assumes that each document is a mixture of a pre-specified number of latent variables (`topics'), where each topic is defined by a distribution of words. The model operates as a hierarchical graphical model, where for each document, a topic distribution is sampled from a Dirichlet prior, and then individual words are generated by first selecting a topic and then sampling a word from that topic's word distribution. LDA has been found to be empirically useful in scenarios where the bag-of-words assumption holds, as it can uncover the underlying thematic structure without considering word order or context.

In our analysis, we applied LDA to a subset of 211,747 reviews from California-based businesses, representing 4.48\% of the overall dataset. Reviews were divided into three sentiment categories based on user ratings: positive (ratings above 3), negative (ratings below 3), and neutral (ratings equal to 3). Within each sentiment category, reviews were grouped by restaurant, and LDA was applied to extract five topics per restaurant. Each topic was defined by the ten most relevant words based on their contribution to the topic’s word distribution.

The topics generated frequently included food-related words (e.g., `meat,' `fish,' `dry,' and `fresh,'); albeit, these words were not tied to particular aspects of the dining experience. Besides, general sentiment words like `good,' `great,' and `like' were common, although their specific context within the reviews was unclear. Words related to less frequently mentioned aspects, such as \textit{ambiance} or \textit{menu variety}, appeared less consistently across the topics. Given these limitations, we chose to focus on the six key aspects extracted by ChatGPT--\textit{service}, \textit{food quality}, \textit{ambiance}, \textit{wait time}, \textit{price}, and \textit{menu variety}. The LDA output, including the most relevant words for each sentiment group, is summarized in Table \ref{lda}. 

To construct sentiment classification models for each identified aspect, we created a labeled dataset for training and evaluation. We started by randomly selecting 5,000 reviews from the full dataset of 4,724,684 reviews, including 1,899 5-star reviews (38.0\%), 1,228 4-star reviews (24.6\%), 657 3-star reviews (13.1\%), 506 2-star reviews (10.1\%), and 710 1-star reviews (14.2\%).  Five annotators independently scored the sentiment of 1,000 randomly assigned reviews for each aspect as positive (+1), neutral (0), or negative (-1). When a review did not discuss a particular aspect, it was marked as `irrelevant' (N/A), since not every review addressed all six aspects being evaluated.

To assess the consistency of sentiment evaluations among annotators, we randomly selected 20 reviews from our 5,000-review dataset and had all annotators evaluate each aspect. Table \ref{labeled_data} illustrates sentiment scores assigned by one annotator for four example reviews. We measured inter-annotator agreement using Fleiss' Kappa with the \texttt{statsmodels} package in Python \cite{seabold2010statsmodels}. Fleiss' Kappa \cite{fleiss1971measuring} is a measure of inter-rater agreement for categorical items. It is calculated using the formula:
$$
\kappa=\frac{P_o-P_e}{1-P_e}
$$

Here, $P_o$ represents the proportion of observed agreement among raters, and $P_e$ represents the proportion of agreement expected by chance. Fleiss' Kappa values range from -1 to 1, where $\kappa=1$ represents perfect agreement, $\kappa=0$ indicates agreement no better than chance, and negative values suggest worse-than-chance agreement. Values between 0.61-0.80 are typically considered substantial agreement, while values above 0.80 indicate almost perfect agreement\cite{landis1977application}. \textit{Price} shows nearly perfect agreement ($\kappa = 0.943$), followed by \textit{menu variety} ($\kappa = 0.789$), \textit{food quality} ($\kappa = 0.774$), \textit{service} ($\kappa = 0.727$), and \textit{ambiance/atmosphere} ($\kappa = 0.675$). \textit{Wait time} showed moderate agreement ($\kappa = 0.460$).

As a complementary measure of inter-rater reliability, we also calculated Pearson's correlation coefficient \cite{freedman2009statistical}. The correlation analysis showed high consistency across most aspects, with \textit{price} showing the highest correlation ($\mu = 0.918, p<0.01$) and \textit{wait time} showing the lowest but still significant correlation ($\mu = 0.745, p<0.05$). Hence, both methods consistently indicate strong inter-annotator agreement across most aspects.




\begin{table*}[htbp]
\centering

\small
\begin{tabular}{@{}c@{\hspace{2em}}c@{}}
\begin{tabular}[t]{|@{}p{0.8cm}@{}|p{0.7cm}p{0.7cm}p{0.7cm}p{0.7cm}p{0.7cm}|@{}}
\multicolumn{6}{l}{\textbf{Service} (Mean = 0.855, p < 0.05)} \\
\hline
& A1 & A2 & A3 & A4 & A5 \\
\hline
A1 & 1 & .83\textsuperscript{***} & .95\textsuperscript{***} & .88\textsuperscript{***} & .78\textsuperscript{***} \\
A2 &  & 1 & .91\textsuperscript{***} & .84\textsuperscript{***} & .83\textsuperscript{***} \\
A3 &  &  & 1 & .90\textsuperscript{***} & .85\textsuperscript{***} \\
A4 &  &  &  & 1 & .85\textsuperscript{***} \\
A5 &  &  &  &  & 1 \\
\hline
\end{tabular}
&
\begin{tabular}[t]{|@{}p{0.8cm}@{}|p{0.7cm}p{0.7cm}p{0.7cm}p{0.7cm}p{0.7cm}|@{}}
\multicolumn{6}{l}{\textbf{Food Quality} (Mean = 0.883, p < 0.01)} \\
\hline
& A1 & A2 & A3 & A4 & A5 \\
\hline
A1 & 1 & .90\textsuperscript{***} & .92\textsuperscript{***} & .86\textsuperscript{***} & .82\textsuperscript{***} \\
A2 &  & 1 & .92\textsuperscript{***} & .90\textsuperscript{***} & .92\textsuperscript{***} \\
A3 &  &  & 1 & .81\textsuperscript{***} & .90\textsuperscript{***} \\
A4 &  &  &  & 1 & .88\textsuperscript{***} \\
A5 &  &  &  &  & 1 \\
\hline
\end{tabular}
\\[2em]
\begin{tabular}[t]{|@{}p{0.8cm}@{}|p{0.7cm}p{0.7cm}p{0.7cm}p{0.7cm}p{0.7cm}|@{}}
\multicolumn{6}{l}{\textbf{Ambiance} (Mean = 0.864, p < 0.05)} \\
\hline
& A1 & A2 & A3 & A4 & A5 \\
\hline
A1 & 1 & .90\textsuperscript{***} & .98\textsuperscript{***} & .79\textsuperscript{***} & .98\textsuperscript{***} \\
A2 &  & 1 & .91\textsuperscript{***} & .72\textsuperscript{***} & .96\textsuperscript{***} \\
A3 &  &  & 1 & .72\textsuperscript{***} & .95\textsuperscript{***} \\
A4 &  &  &  & 1 & .75\textsuperscript{***} \\
A5 &  &  &  &  & 1 \\
\hline
\end{tabular}
&
\begin{tabular}[t]{|@{}p{0.8cm}@{}|p{0.7cm}p{0.7cm}p{0.7cm}p{0.7cm}p{0.7cm}|@{}}
\multicolumn{6}{l}{\textbf{Wait Time} (Mean = 0.745, p < 0.05)} \\
\hline
& A1 & A2 & A3 & A4 & A5 \\
\hline
A1 & 1 & .80\textsuperscript{***} & .54\textsuperscript{*} & .80\textsuperscript{***} & .90\textsuperscript{***} \\
A2 &  & 1 & .52\textsuperscript{*} & .85\textsuperscript{***} & .92\textsuperscript{***} \\
A3 &  &  & 1 & .55\textsuperscript{*} & .61\textsuperscript{**} \\
A4 &  &  &  & 1 & .88\textsuperscript{***} \\
A5 &  &  &  &  & 1 \\
\hline
\end{tabular}
\\[2em]
\begin{tabular}[t]{|@{}p{0.8cm}@{}|p{0.7cm}p{0.7cm}p{0.7cm}p{0.7cm}p{0.7cm}|@{}}
\multicolumn{6}{l}{\textbf{Price} (Mean = 0.918, p < 0.01)} \\
\hline
& A1 & A2 & A3 & A4 & A5 \\
\hline
A1 & 1 & .94\textsuperscript{***} & .91\textsuperscript{***} & .99\textsuperscript{***} & .98\textsuperscript{***} \\
A2 &  & 1 & .85\textsuperscript{***} & .91\textsuperscript{***} & .92\textsuperscript{***} \\
A3 &  &  & 1 & .90\textsuperscript{***} & .93\textsuperscript{***} \\
A4 &  &  &  & 1 & .96\textsuperscript{***} \\
A5 &  &  &  &  & 1 \\
\hline
\end{tabular}
&
\begin{tabular}[t]{|@{}p{0.8cm}@{}|p{0.7cm}p{0.7cm}p{0.7cm}p{0.7cm}p{0.7cm}|@{}}
\multicolumn{6}{l}{\textbf{Menu Variety} (Mean = 0.872, p < 0.05)} \\
\hline
& A1 & A2 & A3 & A4 & A5 \\
\hline
A1 & 1 & .90\textsuperscript{***} & .89\textsuperscript{***} & .85\textsuperscript{***} & .96\textsuperscript{***} \\
A2 &  & 1 & .86\textsuperscript{***} & .84\textsuperscript{***} & .88\textsuperscript{***} \\
A3 &  &  & 1 & .82\textsuperscript{***} & .88\textsuperscript{***} \\
A4 &  &  &  & 1 & .90\textsuperscript{***} \\
A5 &  &  &  &  & 1 \\
\hline
\end{tabular}
\\[1em]
\multicolumn{2}{c}{\textsuperscript{*}p < 0.05, \textsuperscript{**}p < 0.01, \textsuperscript{***}p < 0.001}
\end{tabular}
    \caption{Inter-annotator agreement results measured by pairwise Pearson correlations across six dining aspects: \textit{service}, \textit{food quality}, \textit{ambiance}, \textit{wait time}, \textit{price}, and \textit{menu variety}.} \label{pearson}
\end{table*}

\subsection{Text Vectorization}

We partitioned our 5000-review labeled dataset into training (80\%, 4,000 reviews) and validation (20\%, 1,000 reviews) sets. Using the \texttt{nltk} library \cite{nltk}, we removed all punctuation and special characters, converted text to lowercase, eliminated common stopwords, and applied lemmatization to reduce words to their base forms.

We experimented with two text vectorization methods: TF-IDF (Term Frequency-Inverse Document Frequency) and fastText. For TF-IDF, We used \texttt{Scikit-learn}'s TF-IDF framework to transform the corpus into 400-dimensional feature vectors, while retaining default settings for other parameters. To address the inherent class imbalance in the dataset, we generated two sets of TF-IDF features: one preserving the original class distribution and another balanced using the Synthetic Minority Over-sampling Technique (SMOTE) \cite{smote}. SMOTE creates synthetic samples for minority classes by interpolating between existing instances, thus mitigating the impact of class imbalance on model performance.

For the fastText approach, we trained a custom model on our corpus of 4.7 million reviews. We implemented the skip-gram algorithm with character n-grams ranging from sizes 3 to 6, generating 100-dimensional output vectors \cite{fasttext}. To assess the quality of the learned word representations, we conducted nearest-neighbor queries and analyzed the semantic similarity of neighboring words. For each review, we computed the L2-normalized element-wise mean of its constituent word vectors to obtain a fixed-size feature representation. FastText was implemented using Python's \texttt{fastText} package.

\subsection{Classification Models and Architectures for Sentiment Labeling}

We evaluated four established classification models using the aforementioned vectorization methods: \textit{Multinomial Naive Bayes }\cite{kibriya2005multinomial}, \textit{Logistic Regression} \cite{hosmer2013applied}, \textit{Random Forest} \cite{breiman2001random}, and \textit{Support Vector Machine (SVM)} \cite{hearst1998support}. All models were implemented using the \texttt{Scikit-learn} library \cite{pedregosa2011scikit}. We evaluate the models using accuracy as an overall performance metric, and also calculate precision, recall, and F1-score to assess performance for each sentiment class (positive, neutral, negative). In classification tasks, accuracy is defined as the ratio of correctly predicted instances to the total number of instances (Accuracy $=\frac{\text { True Positives+True Negatives }}{\text { Total Number of Instances }}$). For each sentiment class, \textit{precision} measures how many of the predicted instances for that class were correct (Precision $=\frac{\text{True Positives}}{\text{True Positives + False Positives}}$), while \textit{recall} measures how many of the actual instances of that class were correctly identified (Recall $=\frac{\text{True Positives}}{\text{True Positives + False Negatives}}$). The F1-score combines these metrics using the harmonic mean (F1 $=2 \times \frac{\text{Precision} \times \text{Recall}}{\text{Precision + Recall}}$). All metrics were computed using the \texttt{Scikit-learn} library \cite{pedregosa2011scikit}.

Since not all reviews discussed the six aspects, a key design choice is whether and how to include reviews that do not mention the target aspects (marked as irrelevant) in the training process. Based on our selected vectorization and classification methods from the previous step, we consider two reasonable approaches:

\paragraph{One-Stage Classifier}  For each aspect, this model treats reviews not mentioning that aspect as neutral, using a three-point sentiment scale: -1 (negative), 0 (neutral/irrelevant), or 1 (positive). Using the labeled dataset of 5000 reviews, we trained fastText embeddings after preprocessing and split the data into 80\% training and 20\% validation sets. For each aspect (\textit{service}, \textit{food quality}, \textit{ambiance}, \textit{wait time}, \textit{price}, and \textit{menu variety}), a multinomial logistic regression model predicts sentiment probabilities across the three classes.

\paragraph{Two-Stage Classifier} This approach separates relevance and sentiment classification for each aspect:
\begin{itemize}
\item \textbf{Stage 1 (Relevance):}  Using the same dataset of 5000 reviews, we created binary labels (relevant/irrelevant) for each aspect based on whether reviews mentioned that aspect. After training fastText embeddings on the preprocessed data, we developed a binomial logistic regression model to classify aspect relevance using an 80-20 train-validation split.

\item \textbf{Stage 2 (Sentiment):} For each aspect, using only reviews that mentioned it, we trained another fastText embedding model and a multinomial logistic regression classifier to categorize aspect-specific sentiments as negative (-1), neutral (0), or positive (1). To address the class imbalance, random oversampling was applied to the minority class during training for each aspect.
\end{itemize}

Both architectures predict the sentiment of each aspect independently and produce a 6-dimensional sentiment vector for each review, corresponding to the six aspects (\textit{service}, \textit{food quality}, \textit{ambiance}, \textit{wait time}, \textit{price}, and \textit{menu variety}).

To generate predictions on the full Yelp restaurant dataset of 4.7 million reviews, we first processed all reviews through trained fastText embeddings. These embeddings were then input into two classifiers in parallel. The one-stage classifier directly predicts sentiment for each review. The two-stage classifier operates sequentially: in the first stage, it classifies each review as either relevant or irrelevant for each aspect. Only reviews marked as relevant proceed to the second stage, where their sentiment is determined; irrelevant reviews receive a neutral rating.

Due to the different evaluation frameworks of the one-stage and two-stage classifiers, a direct comparison is non-trivial. To enable a fair comparison, we first filter the one-stage classifier’s predictions to include only those reviews relevant to each specific aspect. For example, for the \textit{service} aspect, we identified 664 reviews out of the 1000 validation-set reviews as relevant based on ground-truth labels and filtered the one-stage classifier’s predictions of these reviews. Similarly, the Stage 2 predictions of the two-stage classifier were generated using the same set of relevant reviews, which makes a direct comparison possible.

To evaluate the performance differences between the two classifiers, we used the McNemar test \cite{mcnemar1947note}. This test is suited for cases where two models are applied to the same dataset, and the goal is to assess whether there is a statistically significant difference in their predictions. Let $n_{01}$ represent the number of instances where the one-stage classifier made an incorrect prediction and the two-stage classifier made a correct prediction, and $n_{10} $ represent the number of instances where the one-stage classifier made a correct prediction and the two-stage classifier made an incorrect prediction. The McNemar test evaluates whether the difference between these two counts is significant. The test statistic is computed as:

$$\chi^2 = \frac{(n_{01} - n_{10})^2}{n_{01} + n_{10}}$$

Under the null hypothesis that Stage 2 of the two-stage classifier performs no better than the one-stage classifier, this test statistic follows a chi-square distribution with one degree of freedom. We conducted a one-sided McNemar test using the \texttt{statsmodels} package \cite{seabold2010statsmodels} in Python to evaluate whether Stage 2 of the two-stage classifier significantly outperforms the one-stage classifier ($\alpha = 0.05$). This is a critical design choice for determining the final architecture that we use for conducting ABSA over millions of reviews in the full corpus.

\subsection{Regression Models}

To prepare for regression analysis, we aggregated the review-level predictions at the restaurant level across the full dataset of 52,286 restaurants. For each restaurant, we computed six aspect-specific averages (\textit{service}, \textit{food quality}, \textit{ambiance}, \textit{wait time}, \textit{price}, and \textit{menu variety}) independently using all available reviews, based on predictions from both the one-stage and two-stage classifiers, respectively. These aspect-wise average ratings are continuous variables ranging from -1 to +1.

To account for geographic and cuisine-type variations in the regression models, we included the `state' variable from the original dataset to control for geographic differences, and also created a standardized `cuisine' variable. We first extracted cuisine-related terms from the multi-category `category' field using regular expressions, initially using known cuisine types (e.g., `Italian,' `Chinese,' `Mexican,'). We then normalized variations in naming conventions (e.g., `Italian Restaurant' and `Traditional Italian' were standardized to `Italian'). For restaurants with more than one ambiguous cuisine type, we determined the primary cuisine based on the first listed category, or manually reviewed the ambiguous categories to assign the most representative label. After standardization, we retained the ten most frequent cuisines based on their occurrence: American (28.19\%), Italian (14.81\%), Mexican (8.18\%), Chinese (5.66\%), Japanese (2.94\%), Thai (1.43\%), Mediterranean (1.85\%), Indian (1.83\%), Cajun/Creole (1.76\%), and Vietnamese (1.16\%). All other cuisines, accounting for about 32.15\% of the dataset, were categorized as `Other'. Table \ref{business_data} provides examples from the final aggregated dataset.

\begin{table}[h]
    \centering
        \begin{tabular}{|p{0.7in}|p{0.3in}|p{0.3in}|p{0.3in}|p{0.3in}|p{0.3in}|p{0.3in}|p{0.3in}|p{0.3in}|p{0.5in}|}
            \hline
            \textbf{Business ID} & \textbf{Serv-ice} & \textbf{Qual-ity} & \textbf{Ambi-ance} & \textbf{Wait Time} & \textbf{Price} & \textbf{Menu} & \textbf{Over-all} & \textbf{State} & \textbf{Cuisine} \\ 
            \hline
            I8...WsWA & 0.53 & 0.80 & 0.46 & 0.00 & 0.06 & 0.00 & 4.5 & NV & American \\ \hline
            UJ...RlA & 0.27 & 0.77 & 0.07 & -0.01 & -0.01 & 0.03 & 4.5 & LA & Indian \\ \hline
            2L...dzw & -0.84 & -0.30 & 0.00 & -0.15 & 0.00 & 0.00 & 1.5 & FL & Other \\ \hline
            8T...FMg & 0.25 & 0.50 & 0.06 & -0.01 & 0.00 & 0.01 & 3.5 & PA & Italian \\ \hline
            BF...sQ & 0.19 & 0.80 & 0.00 & -0.01 & 0.01 & 0.03 & 4.0 & FL & Mediter-ranean \\ \hline               
        \end{tabular}
    \caption{Example average predicted ratings for each restaurant across six aspects: \textit{service}, \textit{food quality(quality)}, \textit{ambiance}, \textit{wait time}, \textit{price}, and \textit{menu variety (menu)} The Overall column represents the restaurant's overall rating. Business IDs are partially masked for anonymity, and the \textit{State} and \textit{Cuisine} columns indicate the location and main cuisine type offered, respectively.}
    \label{business_data}
\end{table}

To quantify the relationship between predicted aspect ratings and overall business ratings while accounting for other factors, we applied linear regression analysis \cite{freedman2009statistical} using R. The general regression model is specified as follows:

\begin{equation}
\begin{split}
\text{Overall rating} &= \beta_0 \\
&\quad + \beta_1 \cdot (\text{Avg. service rating}) \\
&\quad + \beta_2 \cdot (\text{Avg. food quality rating}) \\
&\quad + \beta_3 \cdot (\text{Avg. ambiance rating}) \\
&\quad + \beta_4 \cdot (\text{Avg. wait time rating}) \\
&\quad + \beta_5 \cdot (\text{Avg. menu variety rating}) \\
&\quad + \beta_6 \cdot (\text{Avg. price rating}) \\
&\quad + \sum_{j} \gamma_j X_j \\
&\quad + \epsilon \\
\end{split}
\end{equation}

The term $\sum_{j} \gamma_j X_j$ represents different sets of control variables across four regression model specifications:

\begin{itemize} 
    \item Model 1 (Base): No controls ($X_j = 0$); 
    \item Model 2 (Cuisine-Controlled): Controls for cuisine types ($X_j = X_{\text{cuisine}}$); 
    \item Model 3 (State-Controlled): Controls for state ($X_j = X_{\text{state}}$); 
    \item Model 4 (Fully-Controlled): Controls for both cuisine and state ($X_j = X_{\text{cuisine}} + X_{\text{state}}$). \end{itemize}

Here, $\beta_0$ is the intercept, $\beta_1$ to $\beta_6$ represent the expected changes in the overall rating for a one-unit increase in each corresponding aspect rating, holding other variables constant, and $\epsilon$ denotes the error term. All four models were fitted using both the one-stage and two-stage classifier predictions separately. The complete workflow of our methodology, including data preparation, classification architectures, and regression analysis, is summarized in Figure \ref{workflow} in the Appendices. 

\section{Results}\label{results}

\subsection{Aspect Identification and Sentiment Labeling}

\begin{figure}[h]
    \centering
    \includegraphics[width=\textwidth]{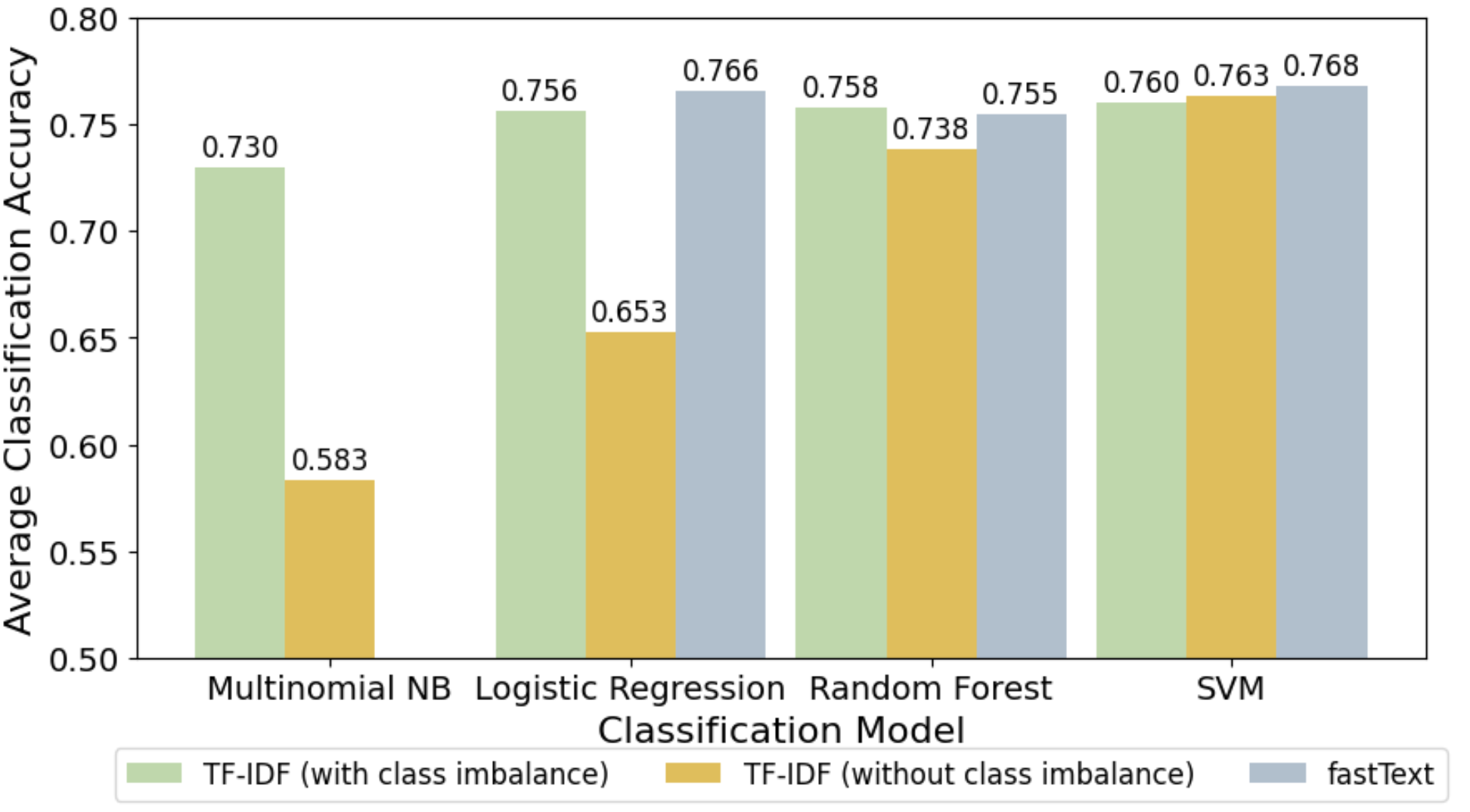} 
    \caption{Average classification accuracy of machine learning algorithms using different text vectorization methods (TF-IDF with and without class imbalance adjustment, and fastText) across all six aspects.}
    \label{model_selection}
\end{figure}

Using the 5,000 labeled reviews, we evaluated four classification models across six attributes --\textit{service}, \textit{food quality}, \textit{ambiance}, \textit{wait time}, \textit{price}, and \textit{menu variety}-- with two text vectorization methods. Figure \ref{model_selection} shows the average classification accuracy of each model across all six aspects, while Tables \ref{model_complete1} and \ref{model_complete2} in the Appendices detail the breakdown of each model’s performance, including precision, recall, F1-scores, and support for each sentiment category within each aspect. Among these models, SVM had the highest average accuracy (0.768) across all methods, followed by logistic regression (0.766) with fastText, random forest (0.758) with TF-IDF (class imbalance adjusted), and naive Bayes, which performed best with TF-IDF (0.730) but was incompatible with fastText.

\begin{figure}[h]
    \centering
    \includegraphics[width=\textwidth]{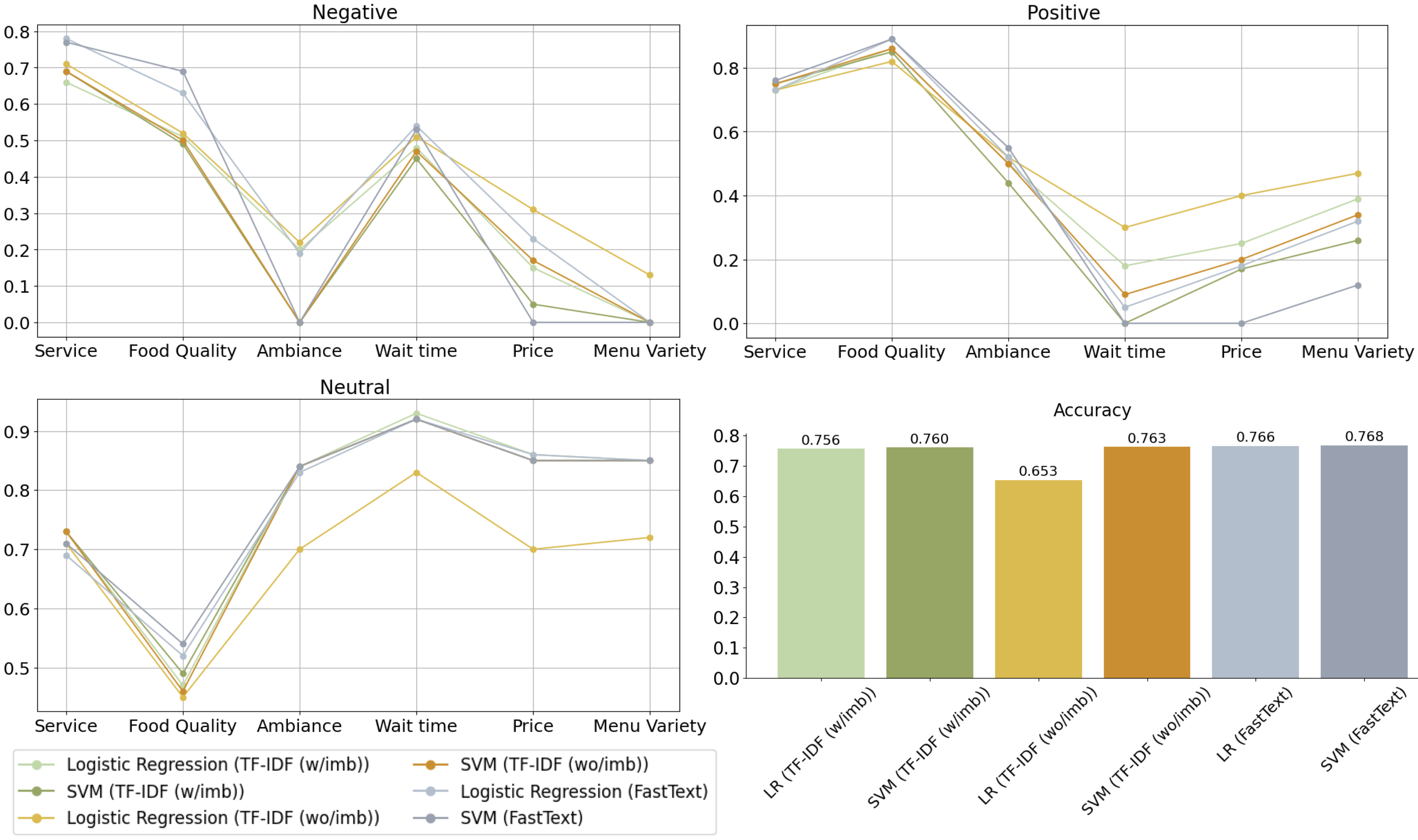} 
    \caption{Performance comparison of machine learning models across restaurant review aspects and sentiment classifications. This figure presents F1-scores and overall accuracy for Support Vector Machines (SVM), Logistic Regression (L.R.), and different vectorization approaches (TF-IDF with/without class imbalance handling, and FastText) across six restaurant aspects. For each aspect, F1-scores are shown for three sentiment categories (Negative, Neutral/Irrelevant, Positive).}
    \label{F1}
\end{figure}

Figure \ref{F1} shows the F1-scores of logistic regression and SVM with two vectorization methods. Despite its high accuracy, SVM showed significant limitations. With TF-IDF (without class imbalance adjustment), SVM failed to predict any negative reviews for \textit{ambiance} and \textit{menu variety} and could not predict positive reviews for \textit{wait time} with the class imbalance adjusted. Using fastText, SVM also missed both positive and negative predictions for \textit{price}, positive predictions for \textit{wait time}, and negative predictions for \textit{ambiance} and \textit{menu variety}. Additionally, SVM was computationally slow on the full dataset (4.7 million reviews) and was even inefficient with the 5,000-sample subset. Logistic regression, while 0.002 points lower in accuracy, only failed to predict negative reviews for \textit{menu variety}. Considering accuracy, robustness, and computational efficiency, we selected logistic regression with fastText as our final model for both one-stage and two-stage classifiers.

Tables \ref{one-stage} and \ref{two-stage} present the performance metrics for the one-stage and two-stage classifiers across six restaurant aspects. On accuracy, the one-stage classifier demonstrates consistent performance, with accuracy ranging from 0.73 for \textit{service} to 0.85 for \textit{wait time}. In the two-stage classifier, Stage 1 achieves strong performance with accuracies between 0.71 for \textit{price} and 0.91 for \textit{food quality}. However, Stage 2, despite being trained exclusively on relevant reviews, generally underperforms the one-stage classifier, with accuracies ranging from 0.61 for \textit{wait time} to 0.76 for \textit{service} and \textit{food quality}. Besides, Stage 2 of the two-stage classifier improves in detecting negative sentiments compared to one-stage classifier, with F1-scores increasing by an average of 0.31 (e.g., from 0.23 to 0.71 for \textit{price} and from 0 to 0.53 for \textit{menu}), but shows significant reduction in F1-scores on neutral class detection, with an average decrease of 0.39 e.g., from 0.86 to 0.42 for \textit{price}, and 0.92 to 0.39 for \textit{wait time}. 

We also observed that the sentiment distribution in some categories is skewed. For instance, \textit{food quality} includes mostly positive reviews (621 positives vs. 160 negatives and 219 neutrals), while \textit{wait time} is predominantly neutral or irrelevant (838 neutrals vs. 78 positives and 84 negatives). Precision and recall appear higher for categories with larger sample sizes in both classifiers—such as positive \textit{food quality}—compared to smaller classes, like negative \textit{menu}, which has only 26 samples. This sample size effect seems more noticeable in the one-stage classifier than in Stage 2 of the two-stage classifier.

As stated earlier in Section \ref{methods}, because the one-stage and two-stage classifiers are based on different premises, it is not straightforward to directly compare their performance. To ensure fair evaluation, we subset the one-stage predictions to include only relevant reviews and compared them with the Stage 2 predictions of the two-stage classifier, which also focused exclusively on relevant reviews. One-sided McNemar tests revealed that the one-stage classifier significantly outperformed Stage 2 across all aspects, even within this controlled comparison. The strongest performance difference was observed for food quality ($\chi^2 = 220.00$, $p < 0.001$), followed by ambiance/atmosphere ($\chi^2 = 117.00$, $p < 0.001$), wait time ($\chi^2 = 83.00$, $p < 0.001$), price ($\chi^2 = 81.00$, $p < 0.001$), and menu variety ($\chi^2 = 73.00$, $p < 0.001$). Even the smallest effect, observed for service ($\chi^2 = 13.36$, $p < 0.001$), remained highly significant.

\begin{table}[h]
    \centering
    \begin{tabular}{|l|c|c|c|c|c|c|}
        \hline
        \multicolumn{7}{|c|}{\textbf{One-Stage Classifier}} \\
        \hline
        \textbf{Metric} & \textbf{Service} & \textbf{Ambiance} & \textbf{Quality} & \textbf{Menu} & \textbf{Wait Time} & \textbf{Price} \\
        \hline
        \multicolumn{7}{|l|}{\textbf{Negative}} \\
        \hline
        Precision & 0.81 & 0.62 & 0.68 & 0 & 0.64 & 0.79 \\
        Recall    & 0.75 & 0.11 & 0.58 & 0 & 0.46 & 0.13  \\
        F1-Score  & 0.78 & 0.19 & 0.63 & 0 & 0.54 & 0.23 \\
        Support  & 191 & 45 & 160 & 26 & 84 & 82 \\
        \hline
        \multicolumn{7}{|l|}{\textbf{Neutral/ Irrelevant}} \\
        \hline
        Precision & 0.67 & 0.78 & 0.60 & 0.77 & 0.87 & 0.76 \\
        Recall    & 0.72 & 0.90 & 0.46 & 0.94 & 0.97 & 0.98 \\
        F1-Score  & 0.69 & 0.83 & 0.52 & 0.85 & 0.92 & 0.86 \\
        Support  & 398 & 692 & 219 & 737 & 838 & 734 \\
        \hline
        \multicolumn{7}{|l|}{\textbf{Positive}} \\
        \hline
        Precision & 0.75 & 0.61 & 0.84 & 0.55 & 0.40 & 0.57 \\
        Recall    & 0.72 & 0.46 & 0.94 & 0.23 & 0.03 & 0.11 \\
        F1-Score  & 0.73 & 0.52 & 0.89 & 0.32 & 0.05 & 0.18 \\
        Support  & 411 & 263 & 621 & 237 & 78 & 184 \\
        \hline
        \textbf{Accuracy}  & 0.73  & 0.74 & 0.78 & 0.75 & 0.85 & 0.75 \\
        \hline
    \end{tabular}
    
    \caption{Performance metrics, including accuracy, class-specific precision, recall, and F1-score, with support of the one-stage classifier across six aspects of restaurant reviews: \textit{service, ambiance, food quality, menu, wait time,} and \textit{price}. }
    \label{one-stage}
\end{table}

\begin{table}[h]
    \centering
    \begin{tabular}{|l|c|c|c|c|c|c|}
        \hline
        \multicolumn{7}{|c|}{\textbf{Two-Stage Classifier}} \\
        \hline
        \textbf{Metric} & \textbf{Service} & \textbf{Ambiance} & \textbf{Quality} & \textbf{Menu} & \textbf{Wait Time} & \textbf{Price} \\
        \hline
        \multicolumn{7}{|c|}{\textbf{Stage 1: Relevance}} \\
        \hline
        \multicolumn{7}{|l|}{\textbf{Relevant}} \\
        \hline
        Precision & 0.76 & 0.75 & 0.91 & 0.69 & 0.73 & 0.82 \\
        Recall    & 0.94 & 0.47 & 1 & 0.12 & 0.17 & 0.06 \\
        F1-Score  & 0.84 & 0.58 & 0.95 & 0.21 & 0.27 & 0.11 \\
        Support  & 664 & 371 & 899 & 296 & 216 & 303 \\
        \hline
        \multicolumn{7}{|l|}{\textbf{Irrelevant}} \\
        \hline
        Precision & 0.77 & 0.75 & 0.86 & 0.73 & 0.81 & 0.71 \\
        Recall    & 0.40 & 0.91 & 0.12 & 0.98 & 0.98 & 0.99 \\
        F1-Score  & 0.53 & 0.82 & 0.21 & 0.83 & 0.89 & 0.83 \\
        Support  & 336 & 629 & 101 & 704 & 784 & 697 \\
        \hline
        \textbf{Accuracy}  & 0.76 & 0.75 & 0.91 & 0.72 & 0.81 & 0.71 \\
        \hline
        \multicolumn{7}{|c|}{\textbf{Stage 2: Sentiment}} \\
        \hline
        \multicolumn{7}{|l|}{\textbf{Negative}} \\
        \hline
        Precision & 0.87 & 0.46 & 0.68 & 0.45 & 0.79 & 0.69 \\
        Recall    & 0.78 & 0.73 & 0.74 & 0.65 & 0.74 & 0.72 \\
        F1-Score  & 0.82 & 0.57 & 0.71 & 0.53 & 0.77 & 0.71 \\
        Support  & 191 & 45 & 160 & 26 & 84 & 82 \\
        \hline
        \multicolumn{7}{|l|}{\textbf{Neutral}} \\
        \hline
        Precision & 0.25 & 0.33 & 0.35 & 0.3 & 0.37 & 0.33 \\
        Recall    & 0.61 & 0.49 & 0.6 & 0.55 & 0.41 & 0.59 \\
        F1-Score  & 0.35 & 0.4 & 0.44 & 0.38 & 0.39 & 0.42 \\
        Support  & 62 & 63 & 118 & 33 & 54 & 37 \\
        \hline
        \multicolumn{7}{|l|}{\textbf{Positive}} \\
        \hline
        Precision & 0.94 & 0.92 & 0.94 & 0.95 & 0.62 & 0.93 \\
        Recall    & 0.78 & 0.72 & 0.79 & 0.79 & 0.62 & 0.77 \\
        F1-Score  & 0.85 & 0.81 & 0.86 & 0.87 & 0.62 & 0.84 \\
        Support  & 411 & 263 & 621 & 237 & 78 & 184 \\
        \hline
        \textbf{Accuracy}  & 0.76 & 0.68 & 0.76 & 0.75 & 0.61 & 0.73 \\
        \hline
    \end{tabular}

\caption{Performance metrics, including accuracy, class-specific precision, recall, F1-score, and support for stage one (relevance classification (relevant vs. irrelevant), as well as sentiment classification (negative, neutral, positive). The two-stage classifier operates across six aspects of restaurant reviews: \textit{service, ambiance, food quality, menu, wait time,} and \textit{price}. Note that Stage 2 was trained exclusively on relevant reviews.}
    
    \label{two-stage}
\end{table}

\subsection{Impact of review aspects on overall restaurant ratings}

\begin{table}[ht]
\centering
\begin{tabular}{lllll}
\toprule
\textbf{Aspect} & \textbf{Model 1} & \textbf{Model 2} & \textbf{Model 3} & \textbf{Model 4} \\
\midrule
Service & 0.75 (0.73, 0.77)*** & 0.74 (0.72, 0.76)*** & 0.78 (0.76, 0.80)*** & 0.77 (0.76, 0.79)*** \\
Food Quality & 1.58 (1.56, 1.60)*** & 1.59 (1.57, 1.60)*** & 1.58 (1.56, 1.60)*** & 1.58 (1.57, 1.60)*** \\
Ambiance & -0.31 (-0.33, -0.29)*** & -0.31 (-0.33, -0.28)*** & -0.27 (-0.30, -0.25)*** & -0.27 (-0.29, -0.25)*** \\
Wait Time & 0.22 (0.15, 0.28)*** & 0.24 (0.17, 0.30)*** & 0.18 (0.12, 0.25)*** & 0.22 (0.15, 0.28)*** \\
Price & 0.06 (-0.05, 0.17) & 0.05 (-0.06, 0.16) & 0.08 (-0.03, 0.19) & 0.07 (-0.04, 0.18) \\
Menu Variety & 0.69 (0.64, 0.74)*** & 0.70 (0.64, 0.75)*** & 0.66 (0.61, 0.72)*** & 0.67 (0.61, 0.72)*** \\
\midrule
$R^2$ & 0.814 & 0.815 & 0.818 & 0.819 \\
\bottomrule
\end{tabular}

\caption{Regression coefficients predicting overall restaurant ratings (one-stage classifier). This table presents the estimated coefficients, 95\% confidence intervals, and significance levels for the six aspects across four regression models. These aspects represent the primary predictors, while control variables (cuisine and state) are analyzed separately in Figure \ref{regression_effect}.  *** indicates significance level of $p<0.001$.}
\label{regression}
\end{table}

To address our second research objective, we applied both the one-stage and two-stage classifiers to predict sentiment labels across all 4.7 million reviews. For each aspect, the one-stage classifier directly predicted sentiment as negative, neutral, or positive. The two-stage classifier first classified reviews as relevant or irrelevant, then predicted sentiment only for the relevant ones. Table \ref{regression} presents regression results based on the one-stage classifier predictions, with the two-stage classifier results shown in Table \ref{two_stage_regression} in the Appendices. Predictions from both classifiers demonstrated strong model fit in the regression analysis, with R-squared values for the one-stage classifier ranging from 0.814 to 0.819.

Across all four models, \textit{food quality} consistently emerges as the strongest predictor of overall ratings, with coefficients ranging from 1.58 to 1.59 (all p-values < 0.001). \textit{Service} follows as the second most important factor, with coefficients between 0.74 and 0.78 (all p-values < 0.05). \textit{Menu variety} also shows a significant positive relationship, with coefficients ranging from 0.66 to 0.70 (all p-values < 0.001). In contrast, \textit{ambiance} has a negative effect, with coefficients between -0.31 and -0.27 (all p-values < 0.001). \textit{Wait time} shows a smaller positive effect, with coefficients between 0.18 and 0.24 (all p-values < 0.001), while \textit{price} does not show statistical significance in any model.

\begin{figure}[h]
    \centering
    \includegraphics[width=\textwidth]{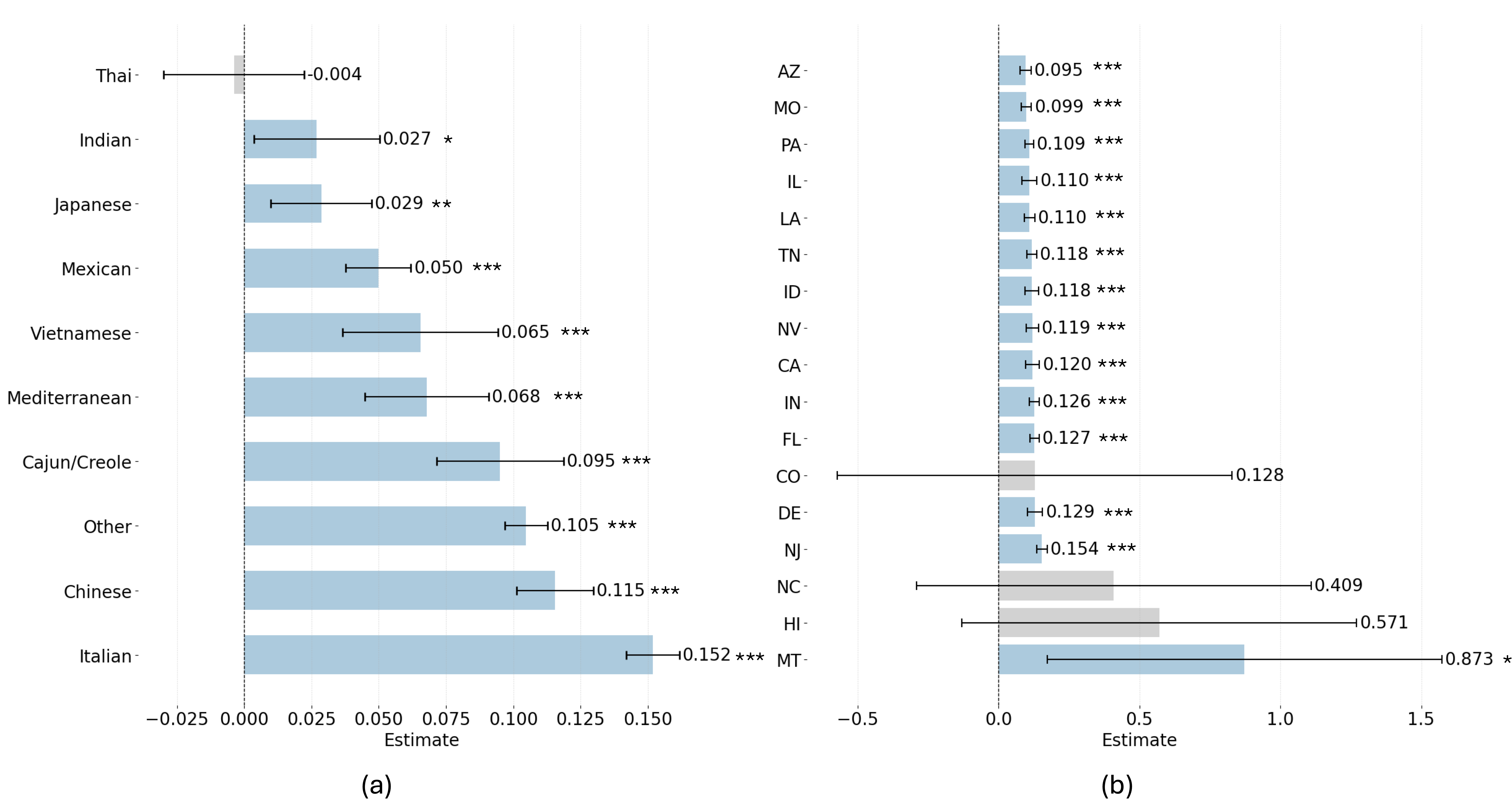} 
    \caption{Visualization of control variable effects from the main regression models. Panel (a) shows the effects of \textit{cuisine} from Model 2, and panel (b) shows the effects of  \textit{state} from Model 3. Horizontal bars represent effect sizes, with 95\% confidence intervals shown as error bars. Categories are ordered by effect magnitude, with numerical estimates displayed alongside each bar. Gray bars indicate non-significant effects ($p>0.05$). Significance levels: *** $p<0.001$, ** $p<0.01$, * $p<0.05$.}
    \label{regression_effect}
\end{figure}

Model 2 suggests that cuisine type is a strong predictor of ratings. As shown in Fig. \ref{regression_effect} (a), using American cuisine as the reference, most cuisines show significant positive associations. Italian cuisine has the strongest positive effect ($\beta$ = 0.152, p < 0.001), followed by Chinese ($\beta$ = 0.115, p < 0.001) cuisine. Thai cuisine is the only type showing a slightly negative effect ($\beta$ = -0.004), though this effect is not statistically significant.

Model 3 reveals significant variation in ratings across different states, suggesting the influence of regional factors. Fig. \ref{regression_effect}(b) indicates that, compared to the reference state (Alberta), Montana shows the largest positive effect ($\beta$ = 0.873, p < 0.05), followed by Hawaii ($\beta$ = 0.571, p > 0.05) and North Carolina ($\beta$ = 0.409, p > 0.05), though the latter two are not statistically significant. Following these, New Jersey ($\beta$ = 0.154, p < 0.001) and Delaware ($\beta$ = 0.129, p < 0.001) show substantial and significant positive effects. The effect sizes for the remaining states are relatively similar, ranging from Arizona ($\beta$ = 0.095) to Florida ($\beta$ = 0.127), all with p < 0.001. Model 4, which includes both cuisine and state controls, yields similar results to Models 2 and 3. Sensitivity analyses using the two-stage classifier confirm these findings. 

\section{Discussion} \label{discussion}

This study addressed two key objectives: (1) using a combination of LLMs and traditional text classification for aspect identification and sentiment labeling at scale, and (2) analyzing the relationship between aspect-based sentiments and overall restaurant ratings. Earlier approaches to customer review analysis primarily relied on frequency-based methods to assess aspects of customer satisfaction \cite{xie2014business, pantelidis2010electronic}. As the volume of online reviews grew, more scalable methods for analysis became necessary. Topic modeling techniques, such as Latent Dirichlet Allocation (LDA) and Latent Semantic Analysis (LSA), became popular tools for discovering latent themes in large datasets without predefined categories. For example, Calheiros et al. \cite{calheiros2017sentiment} applied sentiment classification and LDA to eco-hotel reviews, finding moderate positive sentiments for food and stronger positive feedback for hospitality aspects. Similarly, Akhtar et al. \cite{akhtar2017aspect} used sentiment classification and LDA to support the development of a hotel recommender system. Titov and McDonald \cite{titov2008modeling} introduced topic models for fine-grained analysis of online reviews. Jo and Oh \cite{jo2011aspect} developed the Aspect and Sentiment Unification Model (ASUM) that identifies aspect-sentiment pairs (`senti-aspects') in reviews, achieving sentiment classification comparable to supervised methods without needing labeled data. Xu et al. \cite{xu2016antecedents} used latent semantic analysis on hotel reviews to reveal distinct drivers of satisfaction and dissatisfaction across hotel types.

Advances in machine learning have led researchers to integrate traditional text-mining methods with complex models, including convolutional (CNN) and recurrent neural networks (RNN), for analyzing reviews in the hospitality and tourism sectors. For example, Ma et al. \cite{ma2018effects} used deep learning to assess the informational value of user-provided photos in hotel reviews, showing that combining photos with text improved predictions of review helpfulness compared to using either text or photos alone. Agarwal et al. \cite{agarwal2022deep} applied RNNs and CNNs to classify customer feedback by sentiment and intent, such as complaints or suggestions. Hossain et al. \cite{hossain2020sentiment} used a combined CNN-LSTM model to analyze restaurant reviews, structuring customer feedback data to assess perceptions of restaurant quality, particularly for online delivery services.

More recently, transformer-based models like BERT and GPT have significantly advanced performance on ABSA \cite{devlin2018bert, radford2019language}. These models excel at adapting to word context within sentences, showing improved effectiveness in both aspect identification and sentiment classification, as demonstrated by Xu et al. \cite{xu2019bert} and Sun et al. \cite{sun2019utilizing}. ABSA's applications have also broadened to include fields like product reviews \cite{pontiki2016semeval}, healthcare feedback \cite{bansal2022aspect}, and legal analysis \cite{mudalige2020sigmalaw}. Within hospitality specifically, ABSA has identified key drivers of satisfaction, with food quality, service, and ambiance frequently emerging as top factors \cite{namkung2007does, lin2010restaurant}. Ryu and Han \cite{ryu2010influence} emphasized ambiance's role in loyalty, while Namkung and Jang \cite{namkung2007does} demonstrated food quality's impact on return intentions. Despite these advances, applying ABSA to large, unstructured datasets remains challenging due to noisy and ambiguous data \cite{liu2020aspect}. Hybrid models combining rule-based and deep learning approaches show promise in improving sentiment classification accuracy for complex reviews \cite{ray2022mixed}.
 
In continuing this line of study while addressing our first objective, we explored whether large language models (LLMs) like ChatGPT can complement ABSA by generating refined aspect labels and sentiment scores when labeled data is sparse or inconsistent. Using ChatGPT, we identified six primary aspects (\textit{service}, \textit{food quality}, \textit{ambiance}, \textit{wait time}, \textit{price}, and \textit{menu variety}) across 600 sampled reviews, which are consistent with factors commonly discussed in the literature \cite{zhang2016dimensions, pantelidis2010electronic}. The strong agreement between the two ChatGPT models supports the validity of this approach. To address the scalability challenge outlined in our introduction, we created a human-labeled dataset of 5,000 reviews, which was then used to train a classifier applied to 4.7 million reviews. The strong annotator agreement analyses suggest that the machine-extracted themes were well-defined and consistently identifiable by human annotators.

Addressing our second objective regarding the impact on restaurant ratings, our regression models demonstrated strong goodness-of-fit, with $R^2$ values above 0.8 across all model specifications. This indicates that the machine-labeled aspects can explain a substantial portion of the variance in restaurant ratings and other included variables, though high $R^2$ values do not imply causation. Consistent with prior studies of a smaller-scale or more qualitative nature \cite{namkung2007does,barber2011restaurant}, food quality and service were the strongest predictors of restaurant ratings. We also found a small positive relationship between wait time and ratings, suggesting that longer waits may signal high demand rather than dissatisfaction, potentially supporting the `social proof' theory where customers associate crowded venues with higher quality \cite{cai2009observational}. Ambiance, unexpectedly, showed a negative association with ratings, contrasting with previous findings such as Ryu and Han \cite{ryu2010influence}. We, however, also note moderate annotator agreement ($\kappa = 0.675$), indicating some variability in labeled perceptions of ambiance, which may have influenced the model’s predictions. Future investigation is also warranted.

We also found stronger positive correlations between Italian and Chinese cuisines and higher ratings than American cuisine. Regional differences in ratings were observed, likely influenced by local preferences, economic conditions, and market competition. While research on regional variations in restaurant ratings is limited, broader studies suggest market conditions significantly influence customer preferences \cite{parsa2011restaurants}. The high estimates observed in Montana may be attributed to the limited sample size (only six data points).

This study has several limitations. First, we focused on only six LLM-identified aspects, which may overlook other dimensions of customer experience, especially since our initial analysis was based on just 600 reviews. Additionally, potential biases from manual labeling or the limited geographic coverage of the Yelp dataset across states cannot be entirely ruled out despite the high R-squared values observed in our regression models. Finally, while the one-stage classifier was effective for neutral/irrelevant classes, it struggled with detecting negative sentiments. The two-stage classifier performed better with negative sentiments but had limitations in identifying neutral classes. Since the models predict sentiments for each aspect separately, a hybrid model could be considered, which applies the classifier that performs better for each sentiment class and aspect rather than using either the one-stage or two-stage classifier uniformly across all sentiment classes and aspects.

\section{Conclusion and Future Directions}

This study demonstrates the effectiveness of combining large language models with traditional machine learning approaches for analyzing customer feedback at scale in the restaurant business. We find that ChatGPT can effectively identify key aspects from customer reviews, while trained classifiers can reliably scale sentiment analysis across millions of reviews. The strong correlation between predicted aspect sentiments and overall restaurant ratings validates this approach, suggesting that automated methods can extract meaningful patterns from large-scale text data that traditionally required intensive qualitative analysis. Through regression analysis, we quantified the relative importance of different aspects and also found patterns such as the positive correlation between wait times and ratings, and regional variations in customer preferences.

Our results suggest several promising directions for future research. First, the proposed architecture could be extended to other review platforms like Google Reviews or TripAdvisor for cross-platform comparison of customer sentiments. Building on studies like Xiang et al. \cite{xiang2017comparative} and Phillips et al. \cite{phillips2017understanding}, researchers could also use the system to develop industry-specific ontologies from large-scale unstructured data. Beyond hospitality, our framework could be adapted for industries such as retail and healthcare, where customer feedback is important in service improvement \cite{meesala2018service,caemmerer2010customer}. Additionally, studying temporal trends in aspect sentiments could reveal shifts in customer preferences and expectations, such as those influenced by events like COVID-19.


{\bf Conflicts of interest:} The authors have no conflicts of interest to declare.

{\bf Acknowledgments:} None.

{\bf Funding:} This work received no specific funding.

{\bf Data availability statement:} The Yelp data used for this study are available at \url{https://www.yelp.com/dataset}

\bibliography{sn-bibliography}

\section*{Appendices}

CONTINUED ON NEXT PAGE

\begin{landscape}
\begin{table}[htbp]
\scriptsize
\setlength{\tabcolsep}{4pt}
\begin{tabular}{ll|rrrrr|rrrrr|rrrrr}
\hline
& & \multicolumn{5}{c|}{\textbf{Service}} & \multicolumn{5}{c|}{\textbf{Food Quality}} & \multicolumn{5}{c}{\textbf{Atmosphere}} \\
Model & Class & Prec & Rec & F1 & Acc & Supp & Prec & Rec & F1 & Acc & Supp & Prec & Rec & F1 & Acc & Supp \\
\hline
MNB+TFIDF(w) & Pos & .66 & .71 & .68 & .65 & 411 & .69 & .98 & .81 & .69 & 621 & .75 & .17 & .28 & .72 & 263 \\
& Neu & .60 & .66 & .63 & & 398 & .62 & .29 & .39 & & 219 & .72 & .98 & .83 & & 692 \\
& Neg & .81 & .50 & .62 & & 191 & .74 & .12 & .21 & & 160 & .00 & .00 & .00 & & 45 \\
\hline
LR+TFIDF(w) & Pos & .74 & .73 & .73 & .71 & 411 & .80 & .92 & .86 & .73 & 621 & .60 & .43 & .50 & .74 & 263 \\
& Neu & .67 & .73 & .73 & & 398 & .56 & .41 & .47 & & 219 & .78 & .90 & .84 & & 692 \\
& Neg & .76 & .59 & .66 & & 191 & .59 & .46 & .51 & & 160 & 1.00 & .11 & .20 & & 45 \\
\hline
RF+TFIDF(w) & Pos & .72 & .82 & .77 & .74 & 411 & .73 & .96 & .83 & .71 & 621 & .61 & .33 & .43 & .74 & 263 \\
& Neu & .75 & .76 & .75 & & 398 & .61 & .34 & .44 & & 219 & .76 & .94 & .84 & & 692 \\
& Neg & .78 & .54 & .64 & & 191 & .62 & .23 & .34 & & 160 & .00 & .00 & .00 & & 45 \\
\hline
SVM+TFIDF(w) & Pos & .76 & .73 & .75 & .73 & 411 & .78 & .95 & .85 & .74 & 621 & .63 & .34 & .44 & .74 & 263 \\
& Neu & .69 & .79 & .73 & & 398 & .63 & .41 & .49 & & 219 & .76 & .94 & .84 & & 692 \\
& Neg & .79 & .62 & .69 & & 191 & .65 & .40 & .49 & & 160 & .00 & .00 & .00 & & 45 \\
\hline
MNB+TFIDF(wo) & Pos & .69 & .68 & .68 & .67 & 411 & .86 & .77 & .82 & .68 & 621 & .44 & .65 & .52 & .52 & 263 \\
& Neu & .63 & .61 & .62 & & 398 & .42 & .47 & .45 & & 219 & .81 & .46 & .59 & & 692 \\
& Neg & .68 & .76 & .72 & & 191 & .49 & .61 & .54 & & 160 & .14 & .67 & .23 & & 45 \\
\hline
LR+TFIDF(wo) & Pos & .76 & .71 & .73 & .72 & 411 & .87 & .77 & .82 & .68 & 621 & .47 & .59 & .52 & .60 & 263 \\
& Neu & .70 & .73 & .71 & & 398 & .42 & .48 & .45 & & 219 & .83 & .61 & .70 & & 692 \\
& Neg & .69 & .73 & .71 & & 191 & .47 & .59 & .52 & & 160 & .14 & .51 & .22 & & 45 \\
\hline
RF+TFIDF(wo) & Pos & .73 & .78 & .76 & .74 & 411 & .79 & .85 & .82 & .70 & 621 & .52 & .42 & .47 & .70 & 263 \\
& Neu & .77 & .74 & .76 & & 398 & .50 & .46 & .48 & & 219 & .76 & .85 & .80 & & 692 \\
& Neg & .66 & .62 & .64 & & 191 & .52 & .42 & .46 & & 160 & .19 & .09 & .12 & & 45 \\
\hline
SVM+TFIDF(wo) & Pos & .76 & .73 & .75 & .73 & 411 & .79 & .93 & .86 & .73 & 621 & .63 & .41 & .50 & .75 & 263 \\
& Neu & .68 & .78 & .73 & & 398 & .57 & .39 & .46 & & 219 & .77 & .92 & .84 & & 692 \\
& Neg & .78 & .62 & .69 & & 191 & .58 & .44 & .50 & & 160 & .00 & .00 & .00 & & 45 \\
\hline
LR+FT & Pos & .75 & .72 & .73 & .73 & 411 & .84 & .94 & .89 & .78 & 621 & .61 & .46 & .52 & .74 & 263 \\
& Neu & .67 & .72 & .69 & & 398 & .60 & .46 & .52 & & 219 & .78 & .90 & .83 & & 692 \\
& Neg & .81 & .75 & .78 & & 191 & .68 & .58 & .63 & & 160 & .62 & .11 & .19 & & 45 \\
\hline
RF+FT & Pos & .73 & .70 & .72 & .70 & 411 & .79 & .95 & .86 & .75 & 621 & .66 & .39 & .49 & .75 & 263 \\
& Neu & .64 & .72 & .68 & & 398 & .61 & .37 & .46 & & 219 & .76 & .93 & .84 & & 692 \\
& Neg & .79 & .66 & .72 & & 191 & .64 & .51 & .57 & & 160 & .50 & .02 & .04 & & 45 \\
\hline
SVM+FT & Pos & .77 & .75 & .76 & .73 & 411 & .83 & .95 & .89 & .79 & 621 & .64 & .48 & .55 & .75 & 263 \\
& Neu & .69 & .73 & .71 & & 398 & .66 & .46 & .54 & & 219 & .78 & .91 & .84 & & 692 \\
& Neg & .80 & .74 & .77 & & 191 & .74 & .64 & .69 & & 160 & .00 & .00 & .00 & & 45 \\
\hline
\end{tabular}
\caption{Complete model evaluation results for sentiment classification. Results include Positive (Pos), Neutral (Neu), and Negative (Neg) sentiment metrics (Precision, Recall, F1, and Accuracy) across all model combinations: TF-IDF (with and without class imbalance) and FastText vectorization, paired with Multinomial Naive Bayes (MNB), Logistic Regression (LR), Random Forest (RF), and Support Vector Machines (SVM). Metrics are presented for the aspects: \textit{Service}, \textit{Food Quality}, and \textit{Atmosphere}. Class distributions are provided in the Support (Supp) column.}

\label{model_complete1}
\end{table}

\begin{table}[htbp]
\scriptsize
\setlength{\tabcolsep}{4pt}
\begin{tabular}{ll|rrrrr|rrrrr|rrrrr}
\hline
& & \multicolumn{5}{c|}{\textbf{Wait Time}} & \multicolumn{5}{c|}{\textbf{Price}} & \multicolumn{5}{c}{\textbf{Menu}} \\
Model & Class & Prec & Rec & F1 & Acc & Supp & Prec & Rec & F1 & Acc & Supp & Prec & Rec & F1 & Acc & Supp \\
\hline
MNB+TFIDF(w) & Pos & .00 & .00 & .00 & .85 & 78 & .00 & .00 & .00 & .73 & 184 & .86 & .03 & .05 & .74 & 237 \\
& Neu & .85 & 1.00 & .92 & & 838 & .73 & 1.00 & .85 & & 734 & .74 & 1.00 & .85 & & 737 \\
& Neg & .81 & .20 & .32 & & 84 & .00 & .00 & .00 & & 82 & .00 & .00 & .00 & & 26 \\
\hline
LR+TFIDF(w) & Pos & .73 & .10 & .18 & .86 & 78 & .58 & .16 & .25 & .75 & 184 & .55 & .30 & .39 & .75 & 237 \\
& Neu & .87 & .98 & .93 & & 838 & .77 & .98 & .86 & & 734 & .79 & .93 & .85 & & 737 \\
& Neg & .67 & .37 & .48 & & 84 & .64 & .09 & .15 & & 82 & .00 & .00 & .00 & & 26 \\
\hline
RF+TFIDF(w) & Pos & .00 & .00 & .00 & .86 & 78 & .67 & .13 & .22 & .75 & 184 & .64 & .16 & .25 & .75 & 237 \\
& Neu & .86 & 1.00 & .93 & & 838 & .75 & .99 & .86 & & 734 & .76 & .97 & .85 & & 737 \\
& Neg & .85 & .35 & .49 & & 84 & .00 & .00 & .00 & & 82 & .00 & .00 & .00 & & 26 \\
\hline
SVM+TFIDF(w) & Pos & .00 & .00 & .00 & .86 & 78 & .67 & .10 & .17 & .74 & 184 & .61 & .17 & .26 & .75 & 237 \\
& Neu & .86 & .99 & .92 & & 838 & .75 & .99 & .85 & & 734 & .76 & .97 & .85 & & 737 \\
& Neg & .75 & .32 & .45 & & 84 & .67 & .02 & .05 & & 82 & .00 & .00 & .00 & & 26 \\
\hline
MNB+TFIDF(wo) & Pos & .18 & .55 & .28 & .65 & 78 & .28 & .57 & .37 & .46 & 184 & .39 & .68 & .49 & .52 & 237 \\
& Neu & .93 & .66 & .77 & & 838 & .79 & .41 & .54 & & 734 & .83 & .46 & .60 & & 737 \\
& Neg & .34 & .71 & .47 & & 84 & .22 & .63 & .32 & & 82 & .07 & .46 & .12 & & 26 \\
\hline
LR+TFIDF(wo) & Pos & .22 & .50 & .30 & .73 & 78 & .35 & .47 & .40 & .58 & 184 & .41 & .55 & .47 & .61 & 237 \\
& Neu & .93 & .75 & .83 & & 838 & .80 & .62 & .70 & & 734 & .82 & .64 & .72 & & 737 \\
& Neg & .41 & .68 & .51 & & 84 & .22 & .50 & .31 & & 82 & .08 & .35 & .13 & & 26 \\
\hline
RF+TFIDF(wo) & Pos & .22 & .08 & .11 & .83 & 78 & .48 & .24 & .32 & .73 & 184 & .48 & .38 & .42 & .73 & 237 \\
& Neu & .88 & .93 & .91 & & 838 & .77 & .91 & .83 & & 734 & .79 & .87 & .83 & & 737 \\
& Neg & .48 & .50 & .49 & & 84 & .35 & .18 & .24 & & 82 & .25 & .04 & .07 & & 26 \\
\hline
SVM+TFIDF(wo) & Pos & .57 & .05 & .09 & .86 & 78 & .67 & .12 & .20 & .75 & 184 & .57 & .24 & .34 & .76 & 237 \\
& Neu & .87 & .98 & .92 & & 838 & .76 & .98 & .85 & & 734 & .78 & .95 & .85 & & 737 \\
& Neg & .60 & .38 & .47 & & 84 & .62 & .10 & .17 & & 82 & .00 & .00 & .00 & & 26 \\
\hline
LR+FT & Pos & .40 & .03 & .05 & .85 & 78 & .57 & .11 & .18 & .75 & 184 & .55 & .23 & .32 & .75 & 237 \\
& Neu & .87 & .97 & .92 & & 838 & .76 & .98 & .86 & & 734 & .77 & .94 & .85 & & 737 \\
& Neg & .64 & .46 & .54 & & 84 & .79 & .13 & .23 & & 82 & .00 & .00 & .00 & & 26 \\
\hline
RF+FT & Pos & .00 & .00 & .00 & .85 & 78 & .40 & .01 & .02 & .74 & 184 & .57 & .09 & .15 & .74 & 237 \\
& Neu & .86 & .98 & .92 & & 838 & .74 & 1.00 & .85 & & 734 & .75 & .98 & .85 & & 737 \\
& Neg & .62 & .36 & .45 & & 84 & 1.00 & .04 & .07 & & 82 & .00 & .00 & .00 & & 26 \\
\hline
SVM+FT & Pos & .00 & .00 & .00 & .86 & 78 & .00 & .00 & .00 & .73 & 184 & .68 & .06 & .12 & .74 & 237 \\
& Neu & .87 & .98 & .92 & & 838 & .73 & 1.00 & .85 & & 734 & .75 & .99 & .85 & & 737 \\
& Neg & .73 & .42 & .53 & & 84 & .00 & .00 & .00 & & 82 & .00 & .00 & .00 & & 26 \\
\hline

\end{tabular}
\caption{Complete model evaluation results for sentiment classification (continued). Performance metrics for \textit{Wait Time, Price}, and \textit{Menu} aspects across all model and vectorization combinations. }
\label{model_complete2}
\end{table}
\end{landscape}

\begin{landscape}
\begin{table}[htbp]
\scriptsize
\setlength{\tabcolsep}{6pt}
\begin{tabular}{lrrrr}
\hline
\textbf{Aspect/Variable} & \textbf{Model 1} & \textbf{Model 2} & \textbf{Model 3} & \textbf{Model 4} \\
\hline
\multicolumn{5}{l}{\textbf{Main Aspects}} \\
Service & 0.587 (0.011)*** & 0.464 (0.010)*** & 0.449 (0.011)*** & 0.450 (0.011)*** \\
Food Quality & 1.315 (0.009)*** & 1.438 (0.009)*** & 1.466 (0.009)*** & 1.465 (0.009)*** \\
Ambiance & -0.379 (0.012)*** & -0.374 (0.012)*** & -0.341 (0.012)*** & -0.336 (0.012)*** \\
Wait Time & 0.626 (0.027)*** & 0.637 (0.026)*** & 0.588 (0.026)*** & 0.594 (0.026)*** \\
Price & 0.142 (0.041)*** & 0.106 (0.040)** & 0.075 (0.039). & 0.065 (0.039). \\
Menu Variety & 0.173 (0.027)*** & 0.079 (0.026)** & 0.056 (0.026)* & 0.059 (0.026)* \\
\hline
\multicolumn{5}{l}{\textbf{State Effects}} \\
Arizona & -- & 0.028 (0.009)** & -- & 0.050 (0.009)*** \\
California & -- & 0.036 (0.012)** & -- & 0.057 (0.012)*** \\
Colorado & -- & -0.044 (0.329) & -- & -0.051 (0.325) \\
Delaware & -- & 0.051 (0.013)*** & -- & 0.065 (0.012)*** \\
Florida & -- & 0.032 (0.008)*** & -- & 0.050 (0.008)*** \\
Hawaii & -- & 0.347 (0.329) & -- & 0.306 (0.325) \\
Idaho & -- & 0.047 (0.011)*** & -- & 0.067 (0.011)*** \\
Illinois & -- & 0.042 (0.013)*** & -- & 0.068 (0.012)*** \\
Indiana & -- & 0.058 (0.008)*** & -- & 0.079 (0.008)*** \\
Louisiana & -- & 0.023 (0.009)** & -- & 0.040 (0.009)*** \\
Missouri & -- & 0.032 (0.008)*** & -- & 0.051 (0.008)*** \\
Montana & -- & 0.280 (0.329) & -- & 0.336 (0.325) \\
North Carolina & -- & 0.580 (0.329). & -- & 0.659 (0.325)* \\
New Jersey & -- & 0.086 (0.009)*** & -- & 0.090 (0.009)*** \\
Nevada & -- & 0.053 (0.011)*** & -- & 0.072 (0.010)*** \\
Pennsylvania & -- & 0.051 (0.007)*** & -- & 0.062 (0.007)*** \\
Tennessee & -- & 0.025 (0.008)** & -- & 0.052 (0.008)*** \\
Other & -- & -0.634 (0.329). & -- & -0.629 (0.325). \\
\hline
\multicolumn{5}{l}{\textbf{Cuisine Effects}} \\
Cajun/Creole & -- & -- & 0.062 (0.011)*** & 0.067 (0.011)*** \\
Chinese & -- & -- & 0.203 (0.007)*** & 0.203 (0.007)*** \\
Indian & -- & -- & 0.047 (0.011)*** & 0.050 (0.011)*** \\
Italian & -- & -- & 0.120 (0.005)*** & 0.119 (0.005)*** \\
Japanese & -- & -- & 0.095 (0.009)*** & 0.098 (0.009)*** \\
Mediterranean & -- & -- & 0.008 (0.011) & 0.011 (0.011) \\
Mexican & -- & -- & 0.015 (0.006)** & 0.014 (0.006)* \\
Other Cuisine & -- & -- & 0.069 (0.004)*** & 0.070 (0.004)*** \\
Thai & -- & -- & 0.034 (0.012)** & 0.036 (0.012)** \\
Vietnamese & -- & -- & 0.129 (0.013)*** & 0.139 (0.014)*** \\
\hline
\multicolumn{5}{l}{\scriptsize Significance levels: *** p < 0.001, ** p < 0.01, * p < 0.05, . p < 0.1} \\
\end{tabular}
\caption{Regression results predicting overall restaurant ratings (two-stage classifier). Coefficients shown with standard errors in parentheses. Model 1: Global regression without controls; Model 2: Including \textit{state} fixed effects; Model 3: Including \textit{cuisine} fixed effects; Model 4: Including both \textit{state} and \textit{cuisine} fixed effects.}
\label{two_stage_regression}
\end{table}
\end{landscape}

\begin{figure}[h]
    \centering
    \includegraphics[width=\textwidth]{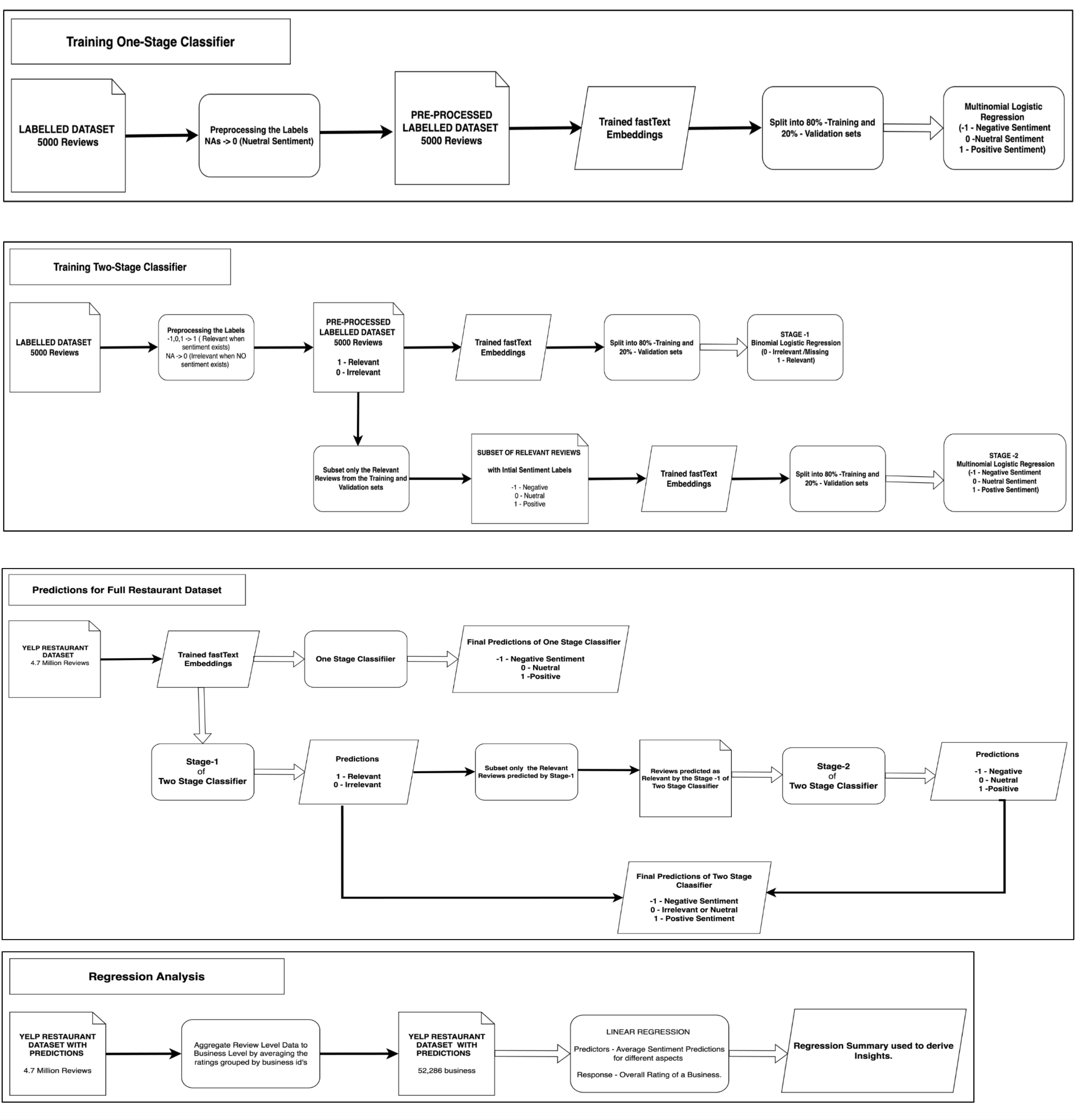} 
    \caption{Workflow visualization for the classification and regression analysis of Yelp restaurant reviews, detailing the training of one-stage and two-stage classifiers, prediction processes based on review relevance and sentiment, and a final regression analysis to model overall business ratings from aggregated sentiment predictions.}

    \label{workflow}
\end{figure}

\end{document}